\documentclass[10pt, a4paper]{article}

\usepackage{lrec-coling2024} 

\usepackage{natbib}
\usepackage{multibib}
\makeatletter
\def\@mb@citenamelist{cite,citep,citet,citealp,citealt,citepalias,citetalias}
\makeatother
\newcites{languageresource}{~}

\usepackage{amsmath}
\usepackage{graphicx}
\usepackage{tabularx}
\usepackage{soul}
\usepackage{enumitem}
\usepackage{makecell}
\usepackage{xcolor}
\usepackage{colortbl}
\usepackage{hyperref}
\usepackage{float}
\usepackage{titlesec}

\usepackage{xcolor}
\usepackage{hyperref}
 \definecolor{darkblue}{rgb}{0, 0, 0.5}
  \hypersetup{colorlinks=true, citecolor=darkblue, linkcolor=darkblue, urlcolor=darkblue}

\usepackage{xstring}

\usepackage{color}

\title{PIRB: A Comprehensive Benchmark of Polish Dense and Hybrid Text Retrieval Methods}

\name{Sławomir Dadas, Michał Perełkiewicz, Rafał Poświata} 

\address{National Information Processing Institute, Warsaw, Poland \\
         \{sdadas, mperelkiewicz, rposwiata\}@opi.org.pl\\}

\abstract{
We present Polish Information Retrieval Benchmark (PIRB), a comprehensive evaluation framework encompassing 41 text information retrieval tasks for Polish. The benchmark incorporates existing datasets as well as 10 new, previously unpublished datasets covering diverse topics such as medicine, law, business, physics, and linguistics. We conduct an extensive evaluation of over 20 dense and sparse retrieval models, including the baseline models trained by us as well as other available Polish and multilingual methods. Finally, we introduce a three-step process for training highly effective language-specific retrievers, consisting of knowledge distillation, supervised fine-tuning, and building sparse-dense hybrid retrievers using a lightweight rescoring model. In order to validate our approach, we train new text encoders for Polish and compare their results with previously evaluated methods. Our dense models outperform the best solutions available to date, and the use of hybrid methods further improves their performance.
 \\ \newline  }

\begin{document}

\maketitleabstract

\section{Introduction}

Text information retrieval is a process of retrieving relevant documents from a large collection of text data in response to a user's query. It is a fundamental task in natural language processing and plays a crucial role in various applications, including search engines, question-answering systems, and recommendation engines. Despite the fact that research in the field of information retrieval has a long history, we have recently observed increased interest in this topic. This is primarily related to the emergence of large language models (LLMs), whether offered as services like GPT-4 \citep{openai2023gpt} or free and publicly available ones like LLaMA \citep{touvron2023llama, touvron2023llama2} or Falcon \citep{penedo2023refinedweb}. With the popularization of these solutions, more attention is also being given to retrieval augmented generation systems \citep{lewis2020retrieval,cai2022recent}, in which a large language model, along with a user's query, receives additional context from an external knowledge base. This context is created based on documents most relevant to the query, extracted using a retrieval algorithm. Correctly selected documents reduce the hallucinations of the language model \citep{shuster2021retrieval} and allow the injection of additional knowledge that the model may not possess. The quality of the response of such a system is therefore highly dependent on the performance of its retrieval component.

\section{Related work}

In recent years, an active area of research has been the use of neural networks, particularly Transformer-based language models, to develop novel text retrieval methods. Many new models have been proposed \citep{yates2021pretrained,zhao2022dense,guo2022semantic,fan2022pre}, leading to significantly improved results on public benchmarks \citep{beir2021} compared to classical term-based approaches such as BM25 \citep{robertson2009probabilistic}. However, this progress has been particularly noticeable for high-resource languages such as English and Chinese, while interest in multilingual and low-resource language solutions has been considerably less pronounced. This can be attributed to the limited availability of datasets, making it challenging to apply the same supervised training methods. This is also related to the lack of standardized benchmarks that would allow for the evaluation of model performance across a wide range of retrieval tasks.

The situation for multilingual retrieval has only recently begun to change with the release of datasets such as mMARCO \citep{bonifacio2021mmarco}, Mr.TyDi \citep{zhang-etal-2021-mr} and MIRACL \citep{zhang2023}, each covering more than 10 languages. In the case of models, multilingual encoders designed for semantic textual similarity \citep{reimers-2020-multilingual-sentence-bert,yang2020multilingual} or for bitext mining \citep{artetxe-schwenk-2019-massively,feng-etal-2022-language} had been available for several years, but there was a lack of high-quality models adapted for text retrieval problems. This improved with the release of the multilingual versions of the E5 \citep{wang2022text} models, which were trained in a weakly supervised manner on a large-scale corpus of text pairs extracted from various online data sources, and subsequently fine-tuned on several manually annotated corpora, including retrieval data.

Despite these efforts, there are still languages not covered in multilingual text retrieval research. One such language is Polish, which has not been featured in any of the aforementioned datasets. To address this issue, the Polish research community has taken the initiative to prepare language-specific corpora. \citet{rybak2022improving} introduced PolQA, a manually annotated dataset for text retrieval containing 7,000 questions matched with passages from Wikipedia. Additionally, some datasets have been created in an automatic or semi-automatic way. \citet{wojtasik2023beir} translated the original BEIR benchmark \citep{beir2021} into Polish using Google Translate and evaluated several baseline retrievers and rerankers. \citet{rybak2023maupqa} made available the MAUPQA collection, comprising almost 400,000 automatically generated, translated, or mined question-answer pairs. Since its original release, the collection has been expanded to include more datasets and currently has over one million text pairs. It is also worth mentioning the PolEval-2022 Passage Retrieval challenge \citep{poleval2022}, which utilized data from the PolQA dataset, and introduced two additional subsets from new domains: legal questions and e-commerce FAQ. 

Although the amount of Polish data for text retrieval is currently sufficiently large, at least compared to low-resource languages, the vast majority of available corpora have been generated automatically, primarily through machine translation. Manually annotated data make up only a small percentage and are limited to a few thousand queries in a single dataset at most. Consequently, some of the generated datasets may be noisy and contain errors resulting from incorrect translations of documents. Therefore, we believe it would still be beneficial to create new datasets containing real questions and answers written or labeled by humans.

In contrast to datasets, there have been few Polish text retrieval models released to date. As part of the work on the MAUPQA collection, two dense retrievers trained on its question-passage pairs have been published \citep{rybak2023maupqa,rybak2023silverretriever}. Furthermore, the authors of the Polish BEIR benchmark provided four rerankers trained by them\footnote{\url{https://huggingface.co/clarin-knext}}.

\section{Contributions}
The aim of our research is to advance Polish text information retrieval in two areas. Our first contribution is to propose a unified benchmark covering a wide range of multi-domain tasks with different characteristics, enabling a reliable and comprehensive evaluation of existing retrieval methods, particularly their generalization ability and zero-shot performance. The second goal is to train and publish new retrieval models for Polish, and then evaluate their performance on the proposed benchmark. The results of our work are described in the article in the following order:

\begin{itemize}[wide,labelwidth=0pt,labelindent=0pt,itemsep=0pt,topsep=5pt]
\item{We present \textbf{Polish Information Retrieval Benchmark (PIRB)} covering 41 text retrieval tasks. The benchmark includes pre-existing dataset collections such as MaupQA, BEIR-PL, and PolEval-2022 Passage Retrieval. We have also prepared 10 new, previously unpublished tasks specifically for the evaluation. Nine of these tasks contain sets of actual questions and answers collected from various Polish websites with diverse topics such as medicine, law, business, physics, or linguistics. The last dataset was generated semi-automatically using GPT-3.5 and includes over 8,000 exam-like questions from over 400 university-level courses.}
\item{We perform an evaluation of more than 20 Polish and multilingual text encoders on the PIRB benchmark. The experiments are performed on already existing dense retrieval models, as well as on strong baseline solutions trained by us, including dense retrievers created using scripts from Sentence-Transformers library and the state-of-the-art sparse retrieval model SPLADE \citep{formal2021splade,Formal2021SPLADEVS}.}
\item{In the last part of the publication, we present our recipe for training highly effective language-specific retrievers. It is a three-step process. First, we use a multilingual knowledge distillation technique \citep{reimers-2020-multilingual-sentence-bert} to transfer knowledge from a high-quality English text encoder to a pre-trained language model for Polish. In the next step, we perform a supervised fine-tuning of the created encoder on the annotated retrieval dataset. The final step is to create a lightweight hybrid retriever, combining the results of the sparse and dense methods using an additional learning-to-rank model. Our proposed technique can be an efficient alternative to solutions that combine the retrieval stage with computationally expensive rerankers. In order to validate the effectiveness of the proposed solution, we apply it to train several Polish text encoders and compare their results with previously evaluated methods on the PIRB benchmark.}
\end{itemize}

Furthermore, we make the developed benchmark publicly available\footnote{\href{https://huggingface.co/spaces/sdadas/pirb}{\tt huggingface.co/spaces/sdadas/pirb}}, as well as the source code of our experiments\footnote{\href{https://github.com/sdadas/pirb}{\tt github.com/sdadas/pirb}}, and the checkpoints of all the models we trained\footnote{\href{https://share.opi.org.pl/s/iS3ziwW9syHdrBj}{\tt share.opi.org.pl/s/iS3ziwW9syHdrBj}}.

\section{Overview of the datasets}
In this section, we present the datasets comprising the PIRB benchmark. We begin with a description of the already existing datasets and dataset collections, and then move on to introduce the corpora we have prepared: web datasets and GPT-exams.

\subsection{Pre-existing datasets}
Of all the datasets included in the PIRB, 31 were previously available as separate benchmarks or individual corpora. When building the benchmark, we sought to collect most of the known publicly available question answering and text retrieval corpora for Polish. We excluded datasets that overlapped with others already found in the benchmark, such as PolQA, whose data was mostly included in queries and passages published for the PolEval-2022 challenge. The following datasets were included in our evaluation:

\begin{itemize}[wide,labelwidth=0pt,labelindent=0pt,itemsep=0pt,topsep=5pt]
\item{\textbf{PolEval-2022 Passage Retrieval} was a competition that took place from mid-2022 to early 2023, and its results were summarized during a workshop organized as part of the FedCSIS conference \citep{poleval2022}. During the competition, successive pieces of data were gradually made available, starting from the training (train) and validation (dev-0) splits, which included only data from Wikipedia, to the test sets (test-A and test-B), each containing three subsets from different domains: Wikipedia, legal questions, and e-commerce FAQ. As part of PIRB, we added 7 data subsets from PolEval: dev-0, as well as three parts each for test-A and test-B.}
\item{\textbf{BEIR-PL} \citep{wojtasik2023beir} aimed to replicate the original BEIR \citep{beir2021} benchmark for Polish. The authors used Google Translate to translate the datasets and published 11 of them, which we included in PIRB.}
\item{\textbf{MAUPQA} \citep{rybak2023maupqa,rybak2023silverretriever} is an ongoing project aimed at assembling a large and diverse corpus of questions and passages that can be used to train Polish retrievers, rerankers, or generative question answering models. The data was mostly collected automatically, and the author employed various techniques to build individual subsets, including machine translation, generation using large language models, transcription of game show recordings, and utilizing data available in partially structured sources for question and answer mining. Some parts of the MAUPQA collection are exact copies of datasets already present in BEIR-PL, so not all of them were included in PIRB. We selected 12 datasets that were added to the benchmark.}
\item{\textbf{MFAQ} \citep{de-bruyn-etal-2021-mfaq} is a multilingual dataset of questions and answers extracted from FAQ pages found in Common Crawl. While the original collection covered 21 languages, in our evaluation we used only the Polish corpus, consisting of more than 60,000 text pairs. }
\end{itemize}

\subsection{Web datasets}
\begin{table*}
\small
\centering
\setlength{\tabcolsep}{7pt}
\renewcommand{\arraystretch}{1.05}
\begin{tabular}{l|l|cc|c|cc}
\hline
\makecell{\textbf{Dataset name}} & \makecell{\textbf{Domain}} & \makecell{\textbf{Total} \\ \textbf{queries}}  & \makecell{\textbf{Avg. words} \\ \textbf{per query}} & \makecell{\textbf{Avg. answers} \\ \textbf{per query}} & \makecell{\textbf{Total} \\ \textbf{documents}}  & \makecell{\textbf{Avg. words} \\ \textbf{per document}} \\
\hline
abczdrowie & medicine & 224,533 & 82 & 1.22 & 274,687 & 77 \\
e-prawnik & law & 35,994 & 63 & 1.00 & 35,994 & 207 \\
gemini & medicine & 272 & 19 & 1.00 & 272 & 120 \\
odi & bussiness & 969 & 8 & 1.60 & 1,546 & 96 \\
onet & trivia & 10,943 & 10 & 1.00 & 10,943 & 27 \\
pwn & linguistics & 16,197 & 41 & 1.00 & 16,197 & 89 \\
specprawnik & law & 54,783 & 66 & 1.22 & 66,832 & 64 \\
techpedia & trivia & 9,205 & 8 & 1.00 & 9,205 & 43 \\
zapytajfizyka & physics & 1,446 & 50 & 1.00 & 1,446 & 211 \\
\hline
\end{tabular}
\caption{\label{tab:dataset_stats}
Characteristics of the web datasets collected by us. For each dataset, we report its domain, number of queries and documents, average query and document length in words, as well as the average number of relevant documents per query.
}
\end{table*}

The PIRB benchmark includes nine datasets crawled from Polish websites. Our goal was to enrich the benchmark with real-world data written originally in Polish. To accomplish this, we identified websites with separate Q\&A sections, some containing questions and answers written directly by the site's editors, and others allowing content creation by registered users. We selected a diverse set of data sources covering various domains such as medicine, law, business, physics, and linguistics. The resulting datasets consist of natural questions, formulated mostly as complete, grammatically correct sentences, as opposed to search engine-oriented datasets such as MS MARCO, in which queries are short and often consist of concatenated keywords. For each data source, we implemented a separate parser to enable precise extraction of question and answer content. We also conducted data cleaning and anonymization, removing personal and contact information. A summary of statistics regarding the collected datasets is presented in Table \ref{tab:dataset_stats}.

The largest datasets originate from websites such as \textbf{abczdrowie} and \textbf{specprawnik}, where content is created by the user community. These are platforms that connect specialists in a particular field, namely medicine and law, with users seeking answers to questions in those fields. Anyone can ask a question, and any registered and verified specialist can provide an answer, allowing these platforms to build large databases of expert advice. The content on these platforms is moderated or redacted by site editors to a minimal extent, resulting in lower data quality compared to other websites we considered. These datasets also required intensive cleaning and filtering. In the resulting data, we included only answers longer than 200 characters and filtered out all unanswered questions. It is also worth noting that in the case of these two datasets, the average length of questions and answers is similar, as users often describe their personal situation with details. Questions have a more individualized character compared to other platforms.

Another group consists of websites that have a dedicated Q\&A section managed by the site staff. Examples of such sites include \textbf{e-prawnik}, \textbf{gemini}, \textbf{odi}, \textbf{pwn}, and \textbf{zapytajfizyka}. Some of these sites allow users to submit their own questions or answers, but these contributions go through editors who review them and decide on their publication. These websites often employ or collaborate with specialists in their respective fields. Datasets created based on these sources are of high quality, but their size is smaller compared to sites relying on user-generated content. Questions are typically shorter and have a general nature, while answers are comprehensive articles providing exhaustive descriptions of the given topic.

The last type of websites includes \textbf{onet} and \textbf{techpedia}. The datasets based on them come from quizzes or collections of trivia questions, the main purpose of which is not to provide advice in a specific field, but primarily to entertain. Users of these sites can complete quizzes to test their knowledge on various topics. Both questions and answers are short. We discarded most data samples from these sources, as they included answers in the form of a single phrase or a short sentence. Only those cases that had a descriptive answer longer than 50 characters were added to the dataset.

\subsection{GPT-exams}
GPT-exams is a dataset created by us in a semi-automatic way utilizing the {\tt gpt-3.5-turbo-0613} model available in the OpenAI API. The dataset contains 8,131 exam-like question-answer pairs covering a wide range of topics. To build the dataset, we performed the following steps:
\begin{enumerate}[wide,labelwidth=0pt,labelindent=0pt,itemsep=0pt,topsep=5pt]
\item{We manually prepared a list of 409 university-level courses from various fields. For each course, we instructed the model with the prompt: "Wygeneruj 20 przykładowych pytań na egzamin z [nazwa przedmiotu]" (Generate 20 sample questions for the [course name] exam). We then parsed the outputs of the model to extract individual questions and performed their deduplication.}
\item{In the next step, we requested the model to generate the answer to each of the collected questions. We used the following prompt: "Odpowiedz na następujące pytanie z dziedziny [nazwa przedmiotu]: [treść pytania]" (Answer the following question from [course name]: [question content]). We sent the following system message with the prompt: "Jesteś ekspertem w dziedzinie [nazwa przedmiotu]. Udzielasz specjalistycznych i wyczerpujących odpowiedzi na pytania." (You are an expert in [course name]. You provide knowledgeable and comprehensive answers to questions).}
\item{In the last step, we manually removed from the dataset the cases in which the model refused to answer the question. We searched for phrases such as "model języka" (language model), "nie jestem" (I'm not), or "nie mogę" (I can't). However, such cases were rare, we found less than 10 refusals for the entire dataset.}
\end{enumerate}

\section{Evaluation}
This section provides a description of the evaluation we conducted on the Polish Information Retrieval Benchmark (PIRB). The experiments include both strong baseline models that we trained as a part of this research and other dense text encoders available for the Polish language. First, we describe the evaluated methods, and then we proceed to discuss the obtained results.

\subsection{Baseline methods}
\begin{table*}
\small
\centering
\setlength{\tabcolsep}{2pt}
\renewcommand{\arraystretch}{1.1}
\begin{tabular}{l|cc|ccccc}
\hline
\textbf{Model name} & \makecell{\textbf{Average} \\ \textbf{NDCG@10} \\ \textbf{(41 tasks)}} & \makecell{\textbf{Avg. without} \\ \textbf{MAUPQA} \\ \textbf{(29 tasks)}} & \makecell{\textbf{PolEval-2022} \\ \textbf{(7 tasks)}} & \makecell{\textbf{Web Datasets} \\ \textbf{(9 tasks)}} & \makecell{\textbf{BEIR-PL} \\ \textbf{(11 tasks)}} & \makecell{\textbf{MAUPQA} \\ \textbf{(12 tasks)}} & \makecell{\textbf{Other} \\ \textbf{(2 tasks)}} \\
\hline
\multicolumn{8}{l}{\textbf{Our sparse baselines}} \\
\hline
BM25 & 41.85 & 43.17 & 45.51 & 47.27 & 33.26 & 38.64 & 71.09 \\
SPLADE++ & 52.93 & 54.06 & 58.92 & 58.60 & 42.47 & \color{purple} \textbf{50.22} & 80.39 \\
\hline
\multicolumn{8}{l}{\textbf{Our dense baselines}} \\
\hline
MSE baseline (base) & 45.47 & 47.82 & 51.87 & 55.56 & 33.55 & 39.80	& 77.35 \\
MSE baseline (large) & 49.98 & 52.47 & 57.49 & 61.13 & 37.26 & 43.98 & 79.58 \\
MNR baseline (base) & 46.44 & 48.98 & 51.69 & 55.71 & 36.52 & 40.30 & 77.74 \\
MNR baseline (large) & 48.63 & 51.18 & 54.22 & 57.82 & 38.61 & 42.45 & 79.84 \\
\hline
\multicolumn{8}{l}{\textbf{Other Polish and multilingual retrievers}} \\
\hline
poleval-2022 (base) & 45.57 & 47.62 & 50.45 & 55.96 & 33.84 & 40.63 & 75.98 \\
poleval-2022 (large) & 48.26 & 50.90 & 53.78 & 58.99 & 37.41 & 41.90 & 78.58 \\
silver-retriever-v1 (base) & $*$ & 53.61 & \color{purple} \textbf{60.87} & \color{purple} \textbf{61.92} & 37.18 & $*$ & \color{purple} \textbf{81.18} \\
multilingual-e5 (small) & 50.65 & 52.26 & 57.84 & 53.82 & 42.45 & 46.77 & 79.72 \\
multilingual-e5 (base) & \color{purple} \textbf{53.12} & \color{purple} \textbf{55.08} & 60.16 & 59.09 & \color{purple} \textbf{44.01} & 48.38 & 80.18 \\
multilingual-e5 (large) & \color{blue} \textbf{57.29} & \color{blue} \textbf{60.09} & \color{blue} \textbf{65.86} & \color{blue} \textbf{64.35} & \color{blue} \textbf{48.99} & \color{blue} \textbf{50.53} & \color{blue} \textbf{81.81} \\
\hline
\end{tabular}
\caption{\label{tab:pirb}
Evaluation results of our baselines and other available retrieval models for Polish. We report Normalized Discounted Cumulative Gain at 10 (NDCG@10) for the entire PIRB benchmark consisting of 41 tasks as well as for separate task groups. The best score in each column is shown in blue, the second best in red. \\ $*$ Since silver-retriever-base-v1 \citep{rybak2023silverretriever} was trained on datasets from MAUPQA, it cannot be reliably evaluated on this part of the benchmark. In order to compare it to the other models, we also provide average scores on the benchmark with MAUPQA datasets excluded.
}
\end{table*}

In our experiments, we included two baseline methods relying on sparse term-based vectors, as well as dense retrievers fine-tuned from Transformer language models. We used the training split of the Polish MS MARCO dataset for all models that required training. The evaluation includes the following sparse approaches:
\begin{itemize}[wide,labelwidth=0pt,labelindent=0pt,itemsep=0pt,topsep=5pt]
\item{\textbf{BM25} \citep{robertson2009probabilistic} is a popular and effective term-based ranking function used in information retrieval since the 1980s. It is an extension of TF-IDF weighting scheme, providing a balanced approach to text ranking, considering both term frequency and document specificity. In our experiments, we used the implementation of BM25 available in anserini \citep{10.1145/3239571}. We applied the default Polish analyzer, which performs the lemmatization of words with  the Morfologik\footnote{\url{https://github.com/morfologik/morfologik-stemming}} library.
\item{\textbf{Sparse Lexical and Expansion (SPLADE)} model \citep{formal2021splade,Formal2021SPLADEVS} is a family of modern term-based methods employing Transformer language models. In this approach, the masked language modeling (MLM) head is optimized to generate a vocabulary-sized weight vector adapted for text retrieval. SPLADE is a highly effective sparse retrieval ranking algorithm, achieving better performance than classic methods such as BM25. Unlike those methods, SPLADE directly uses subwords generated by the tokenizer as terms. We trained the Polish version following the same procedure as the \textbf{SPLADE++} variant \citep{formal2022distillation}, utilizing hard negatives mined with an ensemble of cross-encoders (EnsembleDistil). We used Polish DistilRoBERTa \citep{dadas2020pre} as our base language model.}
}
\end{itemize}

In addition to the above methods, we also trained Polish text encoders using scripts provided in the Sentence-Tranformers library. These examples show how to train high-quality dense retrievers with a set of hard negatives acquired from a cross-encoder. Authors of the library propose two fine-tuning methods with different loss functions:
\begin{itemize}[wide,labelwidth=0pt,labelindent=0pt,itemsep=0pt,topsep=5pt]
\item{
\textbf{MSE (Mean Squared Error) baseline} is a model fine-tuned using the Margin-MSE loss \citep{hofstatter2020improving}, which aims to reduce the margin difference for a positive-negative pair between the reference scores obtained from cross-encoder and scores produced by the optimized model. Specifically, the loss is calculated using the following formula:
\begin{equation}
\mathrm{MSE} \Bigl(s(q, d^{+}) - s(q, d^{-}), \hat{s}(q, d^{+}) - \hat{s}(q, d^{-})\Bigr)
\end{equation}
in which $q$, $d^{+}$ and $d^{-}$ denote query, positive document, and negative document, respectively, $s$ is the reference similarity between query-document pair, and $\hat{s}$ is the similarity computed by the trained model.
}
\item{
\textbf{MNR (Multiple Negatives Ranking) baseline} uses contrastive loss with in-batch negatives \citep{Henderson2017EfficientNL} in addition to hard negatives. Given a query $q_{i}$ and a mini-batch consisting of $K$ queries, the loss for that query is calculated with the following formula:
\begin{equation}
-\hat{s}(q_{i}, d^{+}_{i}) + \log \sum_{\substack{k=1, \\ k \neq i}}^{K} e^{\hat{s}(q_{i}, d^{+}_{k})} + \log \sum_{k=1}^{K} e^{\hat{s}(q_{i}, d^{-}_{k})}
\end{equation}
in which $d^{+}_{k}$ and $d^{-}_{k}$ denote positive and negative document for the $k$-th query, and $\hat{s}$ is the computed text pair similarity.
}
\end{itemize}

We fine-tuned Polish RoBERTa base and large \citep{dadas2020pre} language models with both methods. We used the default hyperparameters defined in the original training scripts, except for the batch size, which we reduced from 64 to 32 for the large models. The models were trained for 10 epochs. We employed a learning rate scheduler with a warmup phase for the first 1,000 batches, a peak learning rate of $2e{-5}$, and linear decay for the remainder of the training. Cosine similarity was used as a similarity metric and mean pooling as the output pooling method.

\subsection{Other methods}
As part of our experiments, we evaluated over 20 publicly available Polish and multilingual dense text encoders, but only a few of them proved to be effective enough for practical use in retrieval tasks. In this publication, we present results only for the models that achieved an NDCG@10 score higher than the BM25 baseline. In the full ranking, which we have made available online\footnote{\href{https://huggingface.co/spaces/sdadas/pirb}{\tt huggingface.co/spaces/sdadas/pirb}}, we include all of the evaluated retrievers, along with more detailed results covering individual datasets. Models offering good performance for Polish text retrieval include:
\begin{itemize}[wide,labelwidth=0pt,labelindent=0pt,itemsep=0pt,topsep=5pt]
\item{\textbf{Multilingual E5} is a text encoder supporting over 100 languages. It is a multilingual version of the E5 model \citep{wang2022text}, developed using the same two-stage training procedure. The first stage involved weakly-supervised training on a dataset of text pairs extracted from large internet corpora, such as Common Crawl. In the second stage, the model was fine-tuned in a supervised manner on several annotated datasets. The multilingual versions, which were released a few months after the original English models, were made available in small (118M parameters), base (278M), and large (560M) sizes.}
\item{\textbf{Silver Retriever} \citep{rybak2023silverretriever} is a Polish dense retrieval model trained on MAUPQA datasets with hard negatives mined employing a combination of heuristic rules and cross-encoders. The model was trained with constrastive loss for 15,000 steps using a batch size of 1024. Only the base-sized version of the encoder was released, fine-tuned from the HerBERT language model \citep{mroczkowski2021herbert}.}
\item{Although the solutions submitted to the \textbf{PolEval-2022} Passage Retrieval competition focused primarily on the reranking phase, a few dense retrievers were also trained. \citet{kozlowski2023} provided two text retrieval models, base and large, fine-tuned on a Polish translation of MS MARCO and training data from the competition, with a set of negatives generated using BM25.
}
\end{itemize}

\subsection{Results}
The results of our evaluation are presented in Table 1. The table compares the performance of our baseline models and the other models described in the previous section. The methods are scored according to the Normalized Discounted Cumulative Gain at 10 (NDCG@10) metric.

We can see that the multilingual E5 models demonstrate high quality, with the large model significantly outperforming the other evaluated solutions. Among the smaller models, in addition to E5 base, Silver Retriever and SPLADE++ also deliver good results. Silver Retriever achieves the second-highest score in three task groups, although worse performance on the BEIR-PL datasets lowers its overall rating. SPLADE++ performs well on all task groups, achieving an average NDCG@10 value close to E5 base, despite being based on a model with only 82 million parameters, fewer than any of the evaluated dense retrievers.

Models from the PolEval-2022 competition and our dense baselines lag behind the other methods. All of these retrievers show similar performance, with MSE large baseline being marginally better than the other models. However, the difference in NDCG@10 between these models and E5 with a similar number of parameters is at least a few points in favor of E5.

\section{Dense and hybrid retrievers}
\begin{figure*}[!ht]
\begin{center}
\includegraphics[scale=0.45]{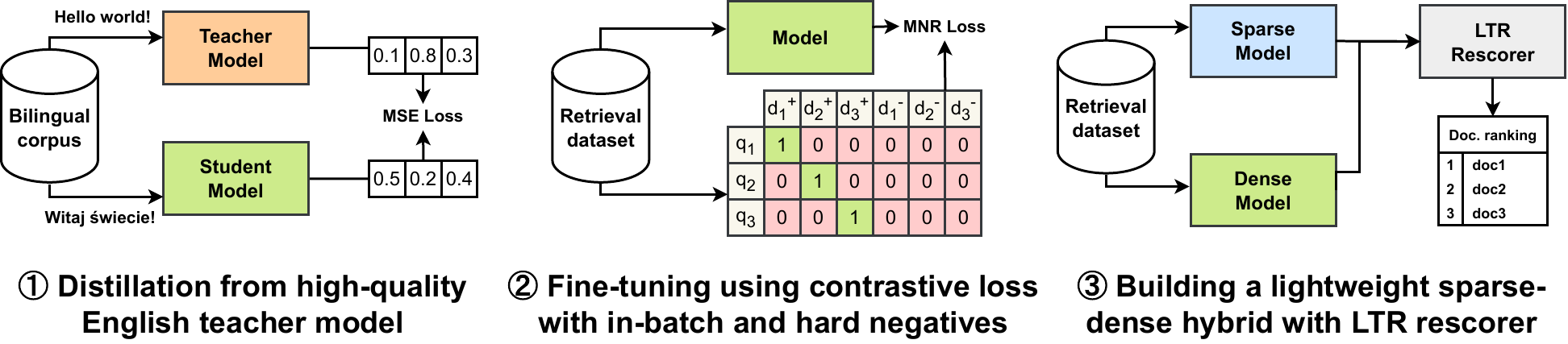} 
\caption{An overview of our procedure for building effective retrieval methods. In the first step, we perform knowledge transfer from an English dense retriever using a bilingual corpus. Next, we fine-tune the obtained model on an annotated dataset for text retrieval in the target language using contrastive loss. In the final step, we construct a lightweight hybrid combining dense and sparse methods, utilizing an additional learning-to-rank model.}
\label{fig:steps}
\end{center}
\end{figure*}

The final stage of our research involved training new dense text encoders for the Polish language that could compete with the best currently available models. In this section, we present our recipe for training effective language-specific retrieval methods. Our proposed process consists of three steps, as illustrated in Figure \ref{fig:steps}. We begin this section by providing an overview of each stage of building our retrieval solution. We then compare the results achieved by our models on the PIRB benchmark with methods tested earlier.

\subsection{Methodology}
Below, we present a description of the individual stages of our experiments. The dense retrieval model resulting from each of the steps is used in the subsequent stage of the process. The order of these steps is as follows:

\begin{enumerate}[wide,labelwidth=0pt,labelindent=0pt,itemsep=0pt,topsep=5pt]
\item{\textbf{Knowledge distillation} - Our procedure begins with transferring knowledge from a high-quality English text encoder to a language-specific model, employing multilingual knowledge distillation technique \citep{reimers-2020-multilingual-sentence-bert}. In this procedure, the original well-performing encoder is known as the teacher model, while the new encoder we want to train is the student model. We initialized the student with the weights of an already trained checkpoint supporting the target language, in our case Polish. We selected two groups of models of different sizes as the basis for our experiments: pre-trained Polish RoBERTa language models \citep{dadas2020pre} and multilingual E5 \citep{wang2022text}. For teachers, we chose English FlagEmbedding \cite{xiao2023cpack} models. The goal of this knowledge distillation method is to fine-tune a student model using bilingual corpora to approximate text representation generated by the teacher. During training, the original encoder generates a vector representation of the English text, while the student produces a representation for the translation of that text. The difference between the vectors is then used to compute the mean-squared error (MSE) loss, which aims to reduce the distance between these representations. 

To conduct this stage of training, we collected a Polish-English bilingual corpus consisting of over 60 million text pairs. The corpus included subsets representing various types of content: 1) single sentences and short texts from the OPUS \citep{tiedemann2012parallel} project; 2) longer paragraph-aligned texts from PELCRA \citep{przepiorkowski2010recent}, DGT \citep{steinberger-etal-2012-dgt}, Wikipedia, and bilingual abstracts from scientific publications; 3) a collection of over 2 million questions extracted from Common Crawl and question-answering datasets, translated from English to Polish using machine translation. The collected texts were subsequently filtered by the LaBSE \citep{feng-etal-2022-language} model to exclude noisy and low-quality translations. We discarded texts for which the similarity between aligned sentences was lower than 0.7. The models were trained with a batch size of 32. We trained small and base models for 3 epochs, large models for 2 epochs. A learning rate scheduler with linear decay and a warmup phase of 10,000 steps was used.
}

\begin{table*}
\small
\centering
\setlength{\tabcolsep}{7pt}
\renewcommand{\arraystretch}{1.1}
\begin{tabular}{l|c|c|c|c|c}
\hline
\textbf{Student model} & \textbf{Teacher model} & \makecell{\textbf{Before} \\ \textbf{fine-tuning}} & \textbf{Fine-tuned} & \makecell{\textbf{Hybrid} \\ \textbf{(BM25)}} & \makecell{\textbf{Hybrid} \\ \textbf{(SPLADE)}} \\
\hline
\multicolumn{6}{l}{\textbf{No distillation, only fine-tuning}} \\
\hline
multilingual-e5 (small) & - & 50.65 & 51.38 (+0.73) & 53.56 (+2.91) & 55.63 (+4.98) \\
multilingual-e5 (base) & - & 53.12 & 54.13 (+1.01) & 55.59 (+2.47) & 56.79 (+3.67) \\
multilingual-e5 (large) & - & \color{blue} \textbf{57.29} & 57.44 (+0.15) & 58.47 (+1.18) & 58.66 (+1.37) \\
\hline
\multicolumn{6}{l}{\textbf{Distillation + fine-tuning}} \\
\hline
multilingual-e5 (small) & bge-small-en & 47.64 & 52.25 (+4.61) & 54.25 (+6.61) & 56.04 (+8.40) \\
multilingual-e5 (base) & bge-base-en & 53.56 & 56.01 (+2.45) & 56.99 (+3.43) & 57.91 (+4.35) \\
multilingual-e5 (large) & bge-large-en & \color{purple} \textbf{56.00} & \color{purple} \textbf{58.03} (+2.03) & \color{blue} \textbf{58.84} (+2.84) & \color{purple} \textbf{59.22} (+3.22) \\
polish-roberta-v2 (base) & bge-base-en & 53.60 & 56.11 (+2.51) & 57.20 (+3.60) & 58.01 (+4.41) \\
polish-roberta-v2 (large) & bge-large-en & 55.62 & \color{blue} \textbf{58.06} (+2.44) & \color{purple} \textbf{58.82} (+3.20) & \color{blue} \textbf{59.23} (+3.61) \\
\hline
\end{tabular}
\caption{\label{tab:retrievers}
A comparison of the NDCG@10 metric on the PIRB benchmark for distilled, fine-tuned and hybrid methods. The best score in each column is shown in blue, the second best in red.
}
\end{table*}

\item{\textbf{Fine-tuning} - The second stage involved further supervised training on an annotated text retrieval dataset. This step is similar to the previously used procedure for training baseline MNR models. In this case, we also used the training split of the MS MARCO dataset, optimizing the models with contrastive loss utilizing in-batch and hard negatives. However, we applied different training hyperparameters. Following \citet{wang2022text}, we introduced a temperature hyperparameter and set it to a value of 0.01. We also used larger batch sizes: 288 for small models, 192 for base models, and 72 for large models. All models were fine-tuned for 50,000 steps, with a warm-up phase for the first 1,000 steps to reach a peak learning rate of $2e-6$, which was then linearly decayed.
}
\item{\textbf{Sparse-dense hybrids} - The final step in our procedure was to build a hybrid system that combines results from dense and sparse models. Since these representations can capture different characteristics of the input data, merging results from two indexes can improve the overall performance of the text retrieval system \citep{chen2022out,luan2021sparse}. Naive aggregation methods involve computing new weights for query-document pairs as a weighted average of the scores returned by each index.  Unfortunately, due to the different scales of the scores, it's difficult to set such weights that would provide good zero-shot results for different datasets. More complex systems use two-stage retrieval, in which the first stage retrieves a number of relevant documents from each index, and then the documents are sorted by the reranker model. However, such systems are considerably slower than purely retrieval-based solutions since the most common reranker architectures require calculating the score for each query-document pair separately.

In our approach, we propose a lightweight learning-to-rank model that takes the scores of dense and sparse models as input and returns a new score for each query-document pair. To train the model, we use the pairwise LambdaMART \citep{burges2010ranknet} algorithm implemented in the XGBoost \citep{chen2016xgboost} library, which optimizes the NDCG metric. For each index that makes up the hybrid system, four attributes are passed to the learning-to-rank model: the score for a given query-document pair, the maximum and minimum score returned for the entire list of relevant documents, and a binary attribute indicating whether a given document is on the relevant documents list. If the document is not present in the set of relevant documents, all four attributes for the index are set to 0. As in the previous stage, we also use a training split of Polish MS MARCO to optimize the rescoring model. We use the following hyperparameters for the XGBRanker model: number of gradient boosted trees is set to 100, max tree depth to 6, subsampling ratio to 0.75, and column subsampling by tree to 0.9. In our experiments, we evaluate two types of hybrid indexes: one using the BM25 method as the sparse index and another utilizing the stronger SPLADE model.

One of the advantages of the proposed rescoring model is its efficiency, resulting in minimal computational overhead on the entire retrieval system. For a hybrid system composed of two indexes and a maximum number of relevant documents set to 200 per query (100 from each index), the throughput of aggregating results using our lightweight learning-to-rank model was approximately 1,500 queries per second on a machine with Intel i7-13700 CPU and Nvidia RTX 3090 GPU. For comparison, using a Transformer-based reranker\footnote{\url{https://huggingface.co/clarin-knext/herbert-base-reranker-msmarco}} on the same configuration achieved a throughput ranging from 2 to 5 queries per second, depending on the dataset and length of the documents in batch.
}
\end{enumerate}

\subsection{Results}
The results of our experiments are presented in Table \ref{tab:retrievers}. We compared the average NDCG@10 values achieved by distilled, fine-tuned, and hybrid models on the PIRB benchmark. We also included an evaluation of the original multilingual E5 models, which exhibited high performance in retrieval tasks. We fine-tuned them on Polish data and built sparse-dense hybrids, just like for the distilled models.

We can observe that after the distillation stage, the quality of the obtained models is lower than that of the available E5 models. An exception is the base-sized models, which already gain a slight advantage at this stage. However, distilled models respond significantly better to fine-tuning. Each model achieves better results after that stage, with distilled models experiencing an increase in the NDCG@10 metric ranging from 2.5 to over 4.5 points. For the original E5 models, this observed increase is lower, at a maximum of 1 point.

The use of hybrid solutions allows for further improvement in results. The observed increases are higher for smaller dense models, for which the combination with a sparse index yields strong results. For instance, the hybrid of the small-sized model with the SPLADE model significantly outperforms standalone base-sized models, even approaching results obtained by large models by around 1 point. For the other model sizes, an increase is also noticeable, with base models gaining an additional 2 points and large models gaining 1 point in NDCG@10 compared to the results from the second stage. As expected, SPLADE hybrids offer better quality than BM25 hybrids, but the difference between these methods also decreases as the size of the dense model increases.

\section{Conclusion}

In this work, we have introduced PIRB, a text information retrieval benchmark for the Polish language consisting of 41 tasks, including 10 that incorporate new, previously unpublished datasets. We conducted a detailed evaluation of multiple retrievers, including dense and hybrid solutions trained as part of our research. The methods built using our approach achieved results surpassing the best text retrieval models for Polish to date. We believe that our work will help standardize the evaluation of information retrieval systems and advance research in this area for Polish.

\section{Acknowledgements}
This research was conducted with the A100 GPU cluster support delivered by the Gdansk University of Technology within the TASK center initiative.

\nocite{*}
\section{Bibliographical References}\label{sec:reference}

\bibliographystyle{lrec-coling2024-natbib}
\bibliography{bibliography}

\begin{thebibliography}{44}
\expandafter\ifx\csname natexlab\endcsname\relax\def\natexlab#1{#1}\fi

\bibitem[{Artetxe and Schwenk(2019)}]{artetxe-schwenk-2019-massively}
Mikel Artetxe and Holger Schwenk. 2019.
\newblock \href {https://doi.org/10.1162/tacl_a_00288} {Massively multilingual sentence embeddings for zero-shot cross-lingual transfer and beyond}.
\newblock \emph{Transactions of the Association for Computational Linguistics}, 7:597--610.

\bibitem[{Bonifacio et~al.(2021)Bonifacio, Jeronymo, Abonizio, Campiotti, Fadaee, , Lotufo, and Nogueira}]{bonifacio2021mmarco}
Luiz~Henrique Bonifacio, Vitor Jeronymo, Hugo~Queiroz Abonizio, Israel Campiotti, Marzieh Fadaee, , Roberto Lotufo, and Rodrigo Nogueira. 2021.
\newblock \href {http://arxiv.org/abs/2108.13897} {mmarco: A multilingual version of ms marco passage ranking dataset}.

\bibitem[{Burges(2010)}]{burges2010ranknet}
Christopher~JC Burges. 2010.
\newblock From {RankNet} to {LambdaRank} to {LambdaMART}: An overview.
\newblock \emph{Learning}, 11(23-581):81.

\bibitem[{Cai et~al.(2022)Cai, Wang, Liu, and Shi}]{cai2022recent}
Deng Cai, Yan Wang, Lemao Liu, and Shuming Shi. 2022.
\newblock Recent advances in retrieval-augmented text generation.
\newblock In \emph{Proceedings of the 45th International ACM SIGIR Conference on Research and Development in Information Retrieval}, pages 3417--3419.

\bibitem[{Chen et~al.(2022)Chen, Zhang, Lu, Bendersky, and Najork}]{chen2022out}
Tao Chen, Mingyang Zhang, Jing Lu, Michael Bendersky, and Marc Najork. 2022.
\newblock Out-of-domain semantics to the rescue! zero-shot hybrid retrieval models.
\newblock In \emph{European Conference on Information Retrieval}, pages 95--110. Springer.

\bibitem[{Chen and Guestrin(2016)}]{chen2016xgboost}
Tianqi Chen and Carlos Guestrin. 2016.
\newblock Xgboost: A scalable tree boosting system.
\newblock In \emph{Proceedings of the 22nd acm sigkdd international conference on knowledge discovery and data mining}, pages 785--794.

\bibitem[{Dadas et~al.(2020)Dadas, Pere{\l}kiewicz, and Po{\'s}wiata}]{dadas2020pre}
S{\l}awomir Dadas, Micha{\l} Pere{\l}kiewicz, and Rafa{\l} Po{\'s}wiata. 2020.
\newblock Pre-training polish transformer-based language models at scale.
\newblock In \emph{Artificial Intelligence and Soft Computing: 19th International Conference, ICAISC 2020, Zakopane, Poland, October 12-14, 2020, Proceedings, Part II 19}, pages 301--314. Springer.

\bibitem[{De~Bruyn et~al.(2021)De~Bruyn, Lotfi, Buhmann, and Daelemans}]{de-bruyn-etal-2021-mfaq}
Maxime De~Bruyn, Ehsan Lotfi, Jeska Buhmann, and Walter Daelemans. 2021.
\newblock \href {https://doi.org/10.18653/v1/2021.mrqa-1.1} {{MFAQ}: a multilingual {FAQ} dataset}.
\newblock In \emph{Proceedings of the 3rd Workshop on Machine Reading for Question Answering}, pages 1--13, Punta Cana, Dominican Republic. Association for Computational Linguistics.

\bibitem[{Fan et~al.(2022)Fan, Xie, Cai, Chen, Ma, Li, Zhang, Guo et~al.}]{fan2022pre}
Yixing Fan, Xiaohui Xie, Yinqiong Cai, Jia Chen, Xinyu Ma, Xiangsheng Li, Ruqing Zhang, Jiafeng Guo, et~al. 2022.
\newblock Pre-training methods in information retrieval.
\newblock \emph{Foundations and Trends{\textregistered} in Information Retrieval}, 16(3):178--317.

\bibitem[{Feng et~al.(2022)Feng, Yang, Cer, Arivazhagan, and Wang}]{feng-etal-2022-language}
Fangxiaoyu Feng, Yinfei Yang, Daniel Cer, Naveen Arivazhagan, and Wei Wang. 2022.
\newblock \href {https://doi.org/10.18653/v1/2022.acl-long.62} {Language-agnostic {BERT} sentence embedding}.
\newblock In \emph{Proceedings of the 60th Annual Meeting of the Association for Computational Linguistics (Volume 1: Long Papers)}, pages 878--891, Dublin, Ireland. Association for Computational Linguistics.

\bibitem[{Formal et~al.(2021{\natexlab{a}})Formal, Lassance, Piwowarski, and Clinchant}]{Formal2021SPLADEVS}
Thibault Formal, C.~Lassance, Benjamin Piwowarski, and St{\'e}phane Clinchant. 2021{\natexlab{a}}.
\newblock \href {https://api.semanticscholar.org/CorpusID:237581550} {Splade v2: Sparse lexical and expansion model for information retrieval}.
\newblock \emph{ArXiv}, abs/2109.10086.

\bibitem[{Formal et~al.(2022)Formal, Lassance, Piwowarski, and Clinchant}]{formal2022distillation}
Thibault Formal, Carlos Lassance, Benjamin Piwowarski, and St{\'e}phane Clinchant. 2022.
\newblock From distillation to hard negative sampling: Making sparse neural ir models more effective.
\newblock In \emph{Proceedings of the 45th International ACM SIGIR Conference on Research and Development in Information Retrieval}, pages 2353--2359.

\bibitem[{Formal et~al.(2021{\natexlab{b}})Formal, Piwowarski, and Clinchant}]{formal2021splade}
Thibault Formal, Benjamin Piwowarski, and St{\'e}phane Clinchant. 2021{\natexlab{b}}.
\newblock Splade: Sparse lexical and expansion model for first stage ranking.
\newblock In \emph{Proceedings of the 44th International ACM SIGIR Conference on Research and Development in Information Retrieval}, pages 2288--2292.

\bibitem[{Guo et~al.(2022)Guo, Cai, Fan, Sun, Zhang, and Cheng}]{guo2022semantic}
Jiafeng Guo, Yinqiong Cai, Yixing Fan, Fei Sun, Ruqing Zhang, and Xueqi Cheng. 2022.
\newblock Semantic models for the first-stage retrieval: A comprehensive review.
\newblock \emph{ACM Transactions on Information Systems (TOIS)}, 40(4):1--42.

\bibitem[{Henderson et~al.(2017)Henderson, Al-Rfou, Strope, Sung, Luk{\'a}cs, Guo, Kumar, Miklos, and Kurzweil}]{Henderson2017EfficientNL}
Matthew Henderson, Rami Al-Rfou, Brian Strope, Yun-Hsuan Sung, L{\'a}szl{\'o} Luk{\'a}cs, Ruiqi Guo, Sanjiv Kumar, Balint Miklos, and Ray Kurzweil. 2017.
\newblock Efficient natural language response suggestion for smart reply.
\newblock \emph{ArXiv}, abs/1705.00652.

\bibitem[{Hofst{\"a}tter et~al.(2020)Hofst{\"a}tter, Althammer, Schr{\"o}der, Sertkan, and Hanbury}]{hofstatter2020improving}
Sebastian Hofst{\"a}tter, Sophia Althammer, Michael Schr{\"o}der, Mete Sertkan, and Allan Hanbury. 2020.
\newblock Improving efficient neural ranking models with cross-architecture knowledge distillation.
\newblock \emph{arXiv preprint arXiv:2010.02666}.

\bibitem[{Kobyliński et~al.(2023)Kobyliński, Ogrodniczuk, Rybak, Przybyła, Pęzik, Mikołajczyk, Janowski, Marcińczuk, and Smywiński-Pohl}]{poleval2022}
Łukasz Kobyliński, Maciej Ogrodniczuk, Piotr Rybak, Piotr Przybyła, Piotr Pęzik, Agnieszka Mikołajczyk, Wojciech Janowski, Michał Marcińczuk, and Aleksander Smywiński-Pohl. 2023.
\newblock {P}ol{E}val 2022/23 challenge tasks and results.
\newblock In \emph{2023 18th Conference on Computer Science and Intelligence Systems (FedCSIS)}, page 1237–1244. IEEE.

\bibitem[{Kozłowski(2023)}]{kozlowski2023}
Marek Kozłowski. 2023.
\newblock Hybrid retrievers with generative re-rankers.
\newblock In \emph{2023 18th Conference on Computer Science and Intelligence Systems (FedCSIS)}, page 1265–1270. IEEE.

\bibitem[{Lewis et~al.(2020)Lewis, Perez, Piktus, Petroni, Karpukhin, Goyal, K{\"u}ttler, Lewis, Yih, Rockt{\"a}schel et~al.}]{lewis2020retrieval}
Patrick Lewis, Ethan Perez, Aleksandra Piktus, Fabio Petroni, Vladimir Karpukhin, Naman Goyal, Heinrich K{\"u}ttler, Mike Lewis, Wen-tau Yih, Tim Rockt{\"a}schel, et~al. 2020.
\newblock Retrieval-augmented generation for knowledge-intensive nlp tasks.
\newblock \emph{Advances in Neural Information Processing Systems}, 33:9459--9474.

\bibitem[{Luan et~al.(2021)Luan, Eisenstein, Toutanova, and Collins}]{luan2021sparse}
Yi~Luan, Jacob Eisenstein, Kristina Toutanova, and Michael Collins. 2021.
\newblock Sparse, dense, and attentional representations for text retrieval.
\newblock \emph{Transactions of the Association for Computational Linguistics}, 9:329--345.

\bibitem[{Mroczkowski et~al.(2021)Mroczkowski, Rybak, Wr{\'o}blewska, and Gawlik}]{mroczkowski2021herbert}
Robert Mroczkowski, Piotr Rybak, Alina Wr{\'o}blewska, and Ireneusz Gawlik. 2021.
\newblock Herbert: Efficiently pretrained transformer-based language model for polish.
\newblock In \emph{Proceedings of the 8th Workshop on Balto-Slavic Natural Language Processing}, pages 1--10.

\bibitem[{OpenAI(2023)}]{openai2023gpt}
R~OpenAI. 2023.
\newblock {GPT}-4 technical report.
\newblock \emph{arXiv}, pages 2303--08774.

\bibitem[{Penedo et~al.(2023)Penedo, Malartic, Hesslow, Cojocaru, Cappelli, Alobeidli, Pannier, Almazrouei, and Launay}]{penedo2023refinedweb}
Guilherme Penedo, Quentin Malartic, Daniel Hesslow, Ruxandra Cojocaru, Alessandro Cappelli, Hamza Alobeidli, Baptiste Pannier, Ebtesam Almazrouei, and Julien Launay. 2023.
\newblock The refinedweb dataset for falcon llm: outperforming curated corpora with web data, and web data only.
\newblock \emph{arXiv preprint arXiv:2306.01116}.

\bibitem[{Przepi{\'o}rkowski et~al.(2010)Przepi{\'o}rkowski, G{\'o}rski, {\L}azinski, and Pezik}]{przepiorkowski2010recent}
Adam Przepi{\'o}rkowski, Rafa{\l}~L G{\'o}rski, Marek {\L}azinski, and Piotr Pezik. 2010.
\newblock Recent developments in the national corpus of polish.
\newblock \emph{NLP, Corpus Linguistics, Corpus Based Grammar Research}, page 302.

\bibitem[{Reimers and Gurevych(2020)}]{reimers-2020-multilingual-sentence-bert}
Nils Reimers and Iryna Gurevych. 2020.
\newblock \href {https://arxiv.org/abs/2004.09813} {Making monolingual sentence embeddings multilingual using knowledge distillation}.
\newblock In \emph{Proceedings of the 2020 Conference on Empirical Methods in Natural Language Processing}. Association for Computational Linguistics.

\bibitem[{Robertson et~al.(2009)Robertson, Zaragoza et~al.}]{robertson2009probabilistic}
Stephen Robertson, Hugo Zaragoza, et~al. 2009.
\newblock The probabilistic relevance framework: Bm25 and beyond.
\newblock \emph{Foundations and Trends{\textregistered} in Information Retrieval}, 3(4):333--389.

\bibitem[{Rybak(2023)}]{rybak2023maupqa}
Piotr Rybak. 2023.
\newblock Maupqa: Massive automatically-created polish question answering dataset.
\newblock In \emph{Proceedings of the 9th Workshop on Slavic Natural Language Processing 2023 (SlavicNLP 2023)}, pages 11--16.

\bibitem[{Rybak and Ogrodniczuk(2023)}]{rybak2023silverretriever}
Piotr Rybak and Maciej Ogrodniczuk. 2023.
\newblock \href {http://arxiv.org/abs/2309.08469} {Silverretriever: Advancing neural passage retrieval for polish question answering}.

\bibitem[{Rybak et~al.(2022)Rybak, Przyby{\l}a, and Ogrodniczuk}]{rybak2022improving}
Piotr Rybak, Piotr Przyby{\l}a, and Maciej Ogrodniczuk. 2022.
\newblock Improving question answering performance through manual annotation: Costs, benefits and strategies.
\newblock \emph{arXiv preprint arXiv:2212.08897}.

\bibitem[{Shuster et~al.(2021)Shuster, Poff, Chen, Kiela, and Weston}]{shuster2021retrieval}
Kurt Shuster, Spencer Poff, Moya Chen, Douwe Kiela, and Jason Weston. 2021.
\newblock Retrieval augmentation reduces hallucination in conversation.
\newblock In \emph{Findings of the Association for Computational Linguistics: EMNLP 2021}, pages 3784--3803.

\bibitem[{Steinberger et~al.(2012)Steinberger, Eisele, Klocek, Pilos, and Schl{\"u}ter}]{steinberger-etal-2012-dgt}
Ralf Steinberger, Andreas Eisele, Szymon Klocek, Spyridon Pilos, and Patrick Schl{\"u}ter. 2012.
\newblock \href {http://www.lrec-conf.org/proceedings/lrec2012/pdf/814_Paper.pdf} {{DGT}-{TM}: A freely available translation memory in 22 languages}.
\newblock In \emph{Proceedings of the Eighth International Conference on Language Resources and Evaluation ({LREC}'12)}, pages 454--459, Istanbul, Turkey. European Language Resources Association (ELRA).

\bibitem[{Thakur et~al.(2021)Thakur, Reimers, R\"{u}ckl\'{e}, Srivastava, and Gurevych}]{beir2021}
Nandan Thakur, Nils Reimers, Andreas R\"{u}ckl\'{e}, Abhishek Srivastava, and Iryna Gurevych. 2021.
\newblock \href {https://datasets-benchmarks-proceedings.neurips.cc/paper_files/paper/2021/file/65b9eea6e1cc6bb9f0cd2a47751a186f-Paper-round2.pdf} {Beir: A heterogeneous benchmark for zero-shot evaluation of information retrieval models}.
\newblock In \emph{Proceedings of the Neural Information Processing Systems Track on Datasets and Benchmarks}, volume~1. Curran.

\bibitem[{Tiedemann(2012)}]{tiedemann2012parallel}
J{\"o}rg Tiedemann. 2012.
\newblock Parallel data, tools and interfaces in opus.
\newblock In \emph{Proceedings of the Eighth International Conference on Language Resources and Evaluation (LREC'12)}, pages 2214--2218.

\bibitem[{Touvron et~al.(2023{\natexlab{a}})Touvron, Lavril, Izacard, Martinet, Lachaux, Lacroix, Rozi{\`e}re, Goyal, Hambro, Azhar et~al.}]{touvron2023llama}
Hugo Touvron, Thibaut Lavril, Gautier Izacard, Xavier Martinet, Marie-Anne Lachaux, Timoth{\'e}e Lacroix, Baptiste Rozi{\`e}re, Naman Goyal, Eric Hambro, Faisal Azhar, et~al. 2023{\natexlab{a}}.
\newblock Llama: Open and efficient foundation language models.
\newblock \emph{arXiv preprint arXiv:2302.13971}.

\bibitem[{Touvron et~al.(2023{\natexlab{b}})Touvron, Martin, Stone, Albert, Almahairi, Babaei, Bashlykov, Batra, Bhargava, Bhosale et~al.}]{touvron2023llama2}
Hugo Touvron, Louis Martin, Kevin Stone, Peter Albert, Amjad Almahairi, Yasmine Babaei, Nikolay Bashlykov, Soumya Batra, Prajjwal Bhargava, Shruti Bhosale, et~al. 2023{\natexlab{b}}.
\newblock Llama 2: Open foundation and fine-tuned chat models.
\newblock \emph{arXiv preprint arXiv:2307.09288}.

\bibitem[{Wang et~al.(2022)Wang, Yang, Huang, Jiao, Yang, Jiang, Majumder, and Wei}]{wang2022text}
Liang Wang, Nan Yang, Xiaolong Huang, Binxing Jiao, Linjun Yang, Daxin Jiang, Rangan Majumder, and Furu Wei. 2022.
\newblock Text embeddings by weakly-supervised contrastive pre-training.
\newblock \emph{arXiv preprint arXiv:2212.03533}.

\bibitem[{Wojtasik et~al.(2023)Wojtasik, Shishkin, Wo{\l}owiec, Janz, and Piasecki}]{wojtasik2023beir}
Konrad Wojtasik, Vadim Shishkin, Kacper Wo{\l}owiec, Arkadiusz Janz, and Maciej Piasecki. 2023.
\newblock Beir-pl: Zero shot information retrieval benchmark for the polish language.
\newblock \emph{arXiv preprint arXiv:2305.19840}.

\bibitem[{Xiao et~al.(2023)Xiao, Liu, Zhang, and Muennighof}]{xiao2023cpack}
Shitao Xiao, Zheng Liu, Peitian Zhang, and Niklas Muennighof. 2023.
\newblock \href {http://arxiv.org/abs/2309.07597} {C-pack: Packaged resources to advance general chinese embedding}.

\bibitem[{Yang et~al.(2018)Yang, Fang, and Lin}]{10.1145/3239571}
Peilin Yang, Hui Fang, and Jimmy Lin. 2018.
\newblock \href {https://doi.org/10.1145/3239571} {Anserini: Reproducible ranking baselines using lucene}.
\newblock \emph{J. Data and Information Quality}, 10(4).

\bibitem[{Yang et~al.(2020)Yang, Cer, Ahmad, Guo, Law, Constant, Abrego, Yuan, Tar, Sung et~al.}]{yang2020multilingual}
Yinfei Yang, Daniel Cer, Amin Ahmad, Mandy Guo, Jax Law, Noah Constant, Gustavo~Hernandez Abrego, Steve Yuan, Chris Tar, Yun-Hsuan Sung, et~al. 2020.
\newblock Multilingual universal sentence encoder for semantic retrieval.
\newblock In \emph{Proceedings of the 58th Annual Meeting of the Association for Computational Linguistics: System Demonstrations}, pages 87--94.

\bibitem[{Yates et~al.(2021)Yates, Nogueira, and Lin}]{yates2021pretrained}
Andrew Yates, Rodrigo Nogueira, and Jimmy Lin. 2021.
\newblock Pretrained transformers for text ranking: Bert and beyond.
\newblock In \emph{Proceedings of the 14th ACM International Conference on web search and data mining}, pages 1154--1156.

\bibitem[{Zhang et~al.(2021)Zhang, Ma, Shi, and Lin}]{zhang-etal-2021-mr}
Xinyu Zhang, Xueguang Ma, Peng Shi, and Jimmy Lin. 2021.
\newblock \href {https://doi.org/10.18653/v1/2021.mrl-1.12} {Mr. {T}y{D}i: A multi-lingual benchmark for dense retrieval}.
\newblock In \emph{Proceedings of the 1st Workshop on Multilingual Representation Learning}, pages 127--137, Punta Cana, Dominican Republic. Association for Computational Linguistics.

\bibitem[{Zhang et~al.(2023)Zhang, Thakur, Ogundepo, Kamalloo, Alfonso-Hermelo, Li, Liu, Rezagholizadeh, and Lin}]{zhang2023}
Xinyu Zhang, Nandan Thakur, Odunayo Ogundepo, Ehsan Kamalloo, David Alfonso-Hermelo, Xiaoguang Li, Qun Liu, Mehdi Rezagholizadeh, and Jimmy Lin. 2023.
\newblock \href {https://doi.org/10.1162/tacl_a_00595} {{MIRACL: A Multilingual Retrieval Dataset Covering 18 Diverse Languages}}.
\newblock \emph{Transactions of the Association for Computational Linguistics}, 11:1114--1131.

\bibitem[{Zhao et~al.(2022)Zhao, Liu, Ren, and Wen}]{zhao2022dense}
Wayne~Xin Zhao, Jing Liu, Ruiyang Ren, and Ji-Rong Wen. 2022.
\newblock Dense text retrieval based on pretrained language models: A survey.
\newblock \emph{arXiv preprint arXiv:2211.14876}.

\end{thebibliography}

\newpage
\appendix
\section{Appendix: Data pre-processing}
This section describes our data pre-processing pipeline for Web Datasets. Since they contain questions and answers collected from the internet, additional cleaning and filtering steps are required to ensure the appropriate quality of evaluation data. For the GPT-exams, there is no need to perform the described steps because the model's responses do not exhibit the same issues. The cleaning process primarily relies on heuristics and dictionary-based methods. More specifically, we employed the following steps:
\begin{itemize}[wide,labelwidth=0pt,labelindent=0pt,itemsep=0pt,topsep=5pt]
\item{\textbf{Text normalization} - We removed e-mails and www addresses using regular expressions. Additionally, special characters were removed, and sequences of whitespace characters were replaced with a single space.}
\item{\textbf{Removing personal information} - In order to anonymize the documents, we split the text into sentences and lemmatized each sentence. Then, we use a set of dictionaries of Polish first names, surnames, and words indicating contact information (such as "mail," "www," "tel", "address") to identify sentences containing such words and remove them from the texts.}
\item{\textbf{Removing non-informative phrases} - Using a dictionary-based method, we also removed phrases from the beginning and end of the text, both from questions and answers. We removed texts such as "good morning", "thank you in advance", "best regards", which do not contribute substantively to the content.}
\item{\textbf{Numbering removal} - In some sources, questions are numbered. For these datasets, we remove the number if it appears at the beginning of the question.}
\item{\textbf{Removing questions with images} - In two datasets (techpedia and onet), there may be questions that contain additional context in the form of an image. In the case of these datasets, we removed all questions containing phrases such as "picture", "photo", "image".}
\item{\textbf{Removing short texts} - Most erroneous and noisy samples come from short questions or answers. Moreover, short answers are often too general to be matched to a specific question without providing additional context. Therefore, we removed questions shorter than 10 characters and answers shorter than 50 characters (200 for abczdrowie and specprawnik, since those two datasets were of lower quality than the rest).}
\end{itemize}

The source code of our preprocessing script, along with the dictionaries and execution arguments for each dataset, is available here: \url{https://share.opi.org.pl/s/oMFJqJ5sSWjzFSX}.

\section{Appendix: Error analysis}
In this section, we present conclusions from error analysis based on the results obtained by our standalone retrieval models and hybrid methods. In order to further investigate the results of our methods, we conducted two comparisons. Below is the summary from the analysis. \\~\\
\noindent \textbf{Best of our dense retrievers (distilled and fine-tuned polish-roberta-large-v2) vs. previous state-of-the-art (multilingual-e5-large):}
\begin{itemize}[wide,labelwidth=0pt,labelindent=0pt,itemsep=0pt,topsep=5pt]
\item{We divided all datasets into subgroups based on average query length: short (less than 10 words), medium (10-20 words), and long (more than 20 words). Our model outperforms E5 on 53\% of short, 50\% of medium, and 83\% of long datasets.}
\item{PolEval-2022 is the only taks group, in which E5 performs significantly better. The group consists of seven datasets, all with a similar structure, comprising short natural queries and relatively short passages (below 50 words).}
\item{On the other hand, our model achieves better results on datasets from the BEIR-PL and Web Datasets groups, which consist of a more diverse sets of tasks. BEIR-PL also includes datasets with short queries from search engines, such as MS MARCO, NFCorpus, or DBPedia. On these datasets, our model has a few points of NDCG@10 advantage over E5.}
\item{E5 performs better or comparably to our model on trivia-like datasets, while our model exhibits better generalization ability to specialized domains.}
\end{itemize}

\noindent \textbf{Our standalone retrievers vs. hybrid retrieval:}
\begin{itemize}[wide,labelwidth=0pt,labelindent=0pt,itemsep=0pt,topsep=5pt]
\item{The use of a hybrid retrieval improves results primarily for queries containing named entities or specialized terminology. Such cases can benefit from term-based matching of sparse indexes.}
\item{In addition, the use of the hybrid approach improves results for task groups, in which the standalone model performed worse or comparably to the original E5 model, particularly for PolEval-2022 and MAUPQA tasks. In the other groups, the differences are small.}
\end{itemize}

\section{Appendix: Detailed results}
\label{sec:appendix-results}
In the main part of the publication, we focused only on the subset of models evaluated by us, which we assessed according to the NDCG@10 metric. Below are more detailed results of our experiments, demonstrating a wider range of metrics including NNDCG@10, MRR@10, Recall@100, and Accuracy@1.

\begin{figure*}[htp]
\includegraphics[scale=0.53]{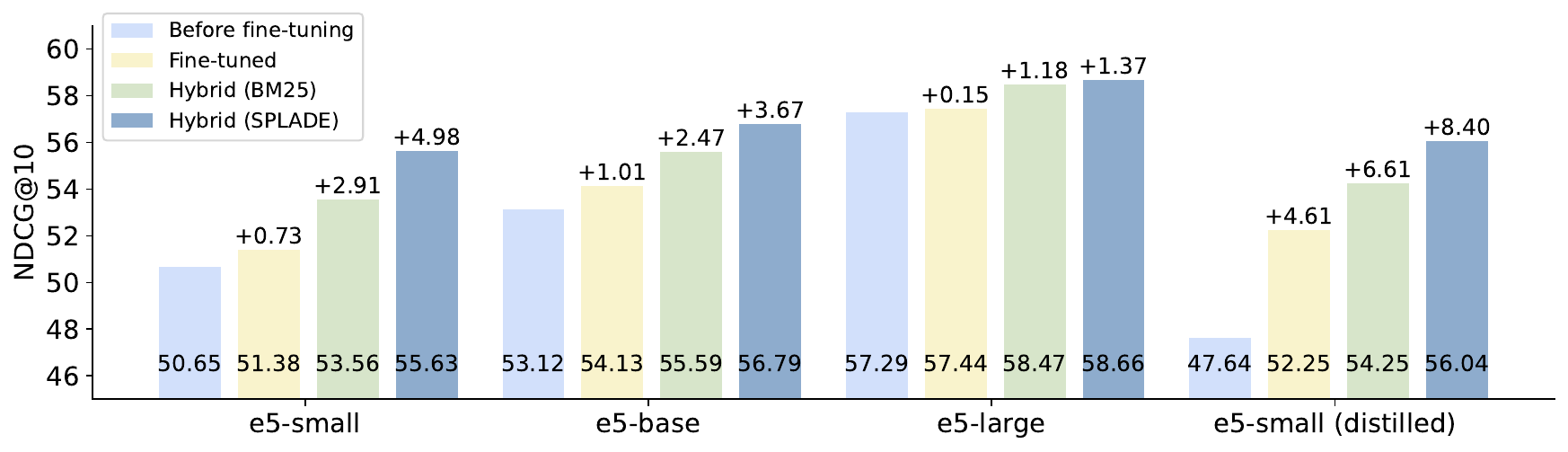}\\
\includegraphics[scale=0.53]{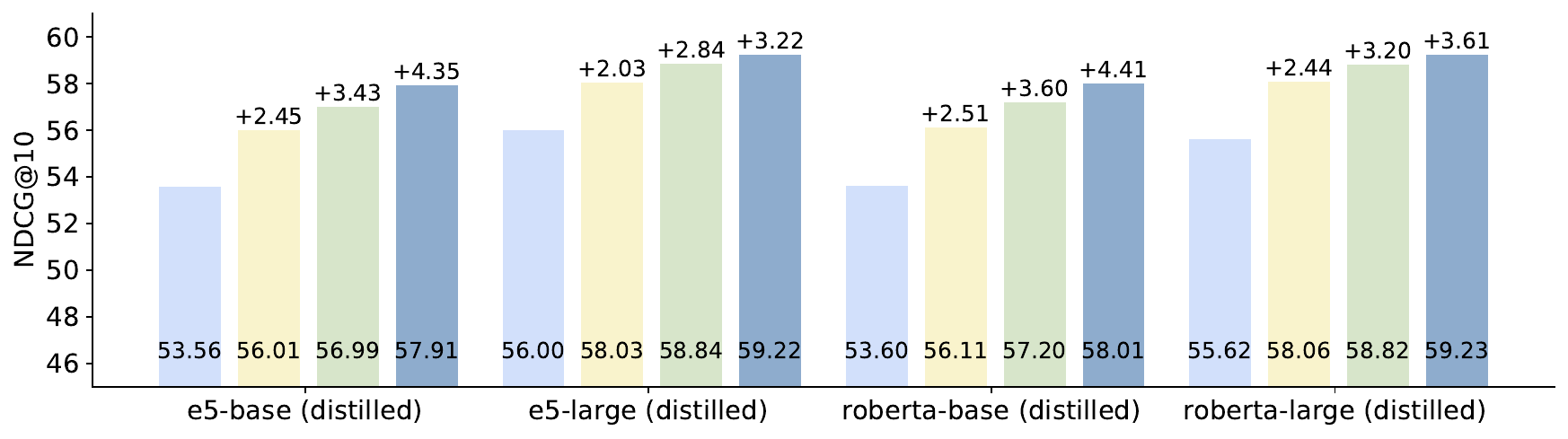}
\caption{NDCG@10 values obtained by the models trained in our study. We present the results of the original multilingual E5 models, as well as models distilled from FlagEmbeddings based on E5 and Polish RoBERTa. For each model, we show its performance before fine-tuning, after fine-tuning on the Polish MS MARCO dataset, and the result of the sparse-dense hybrid combining the given model with BM25 and SPLADE methods.}
\label{fig:ndcg}
\end{figure*}

\begin{figure*}[htp]
\includegraphics[scale=0.53]{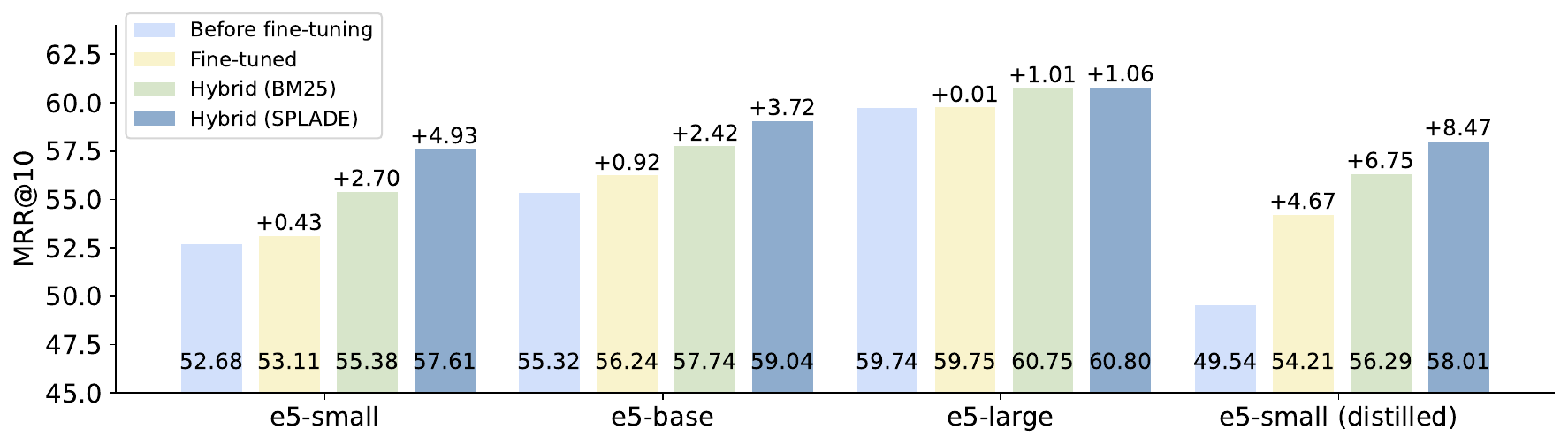}\\
\includegraphics[scale=0.53]{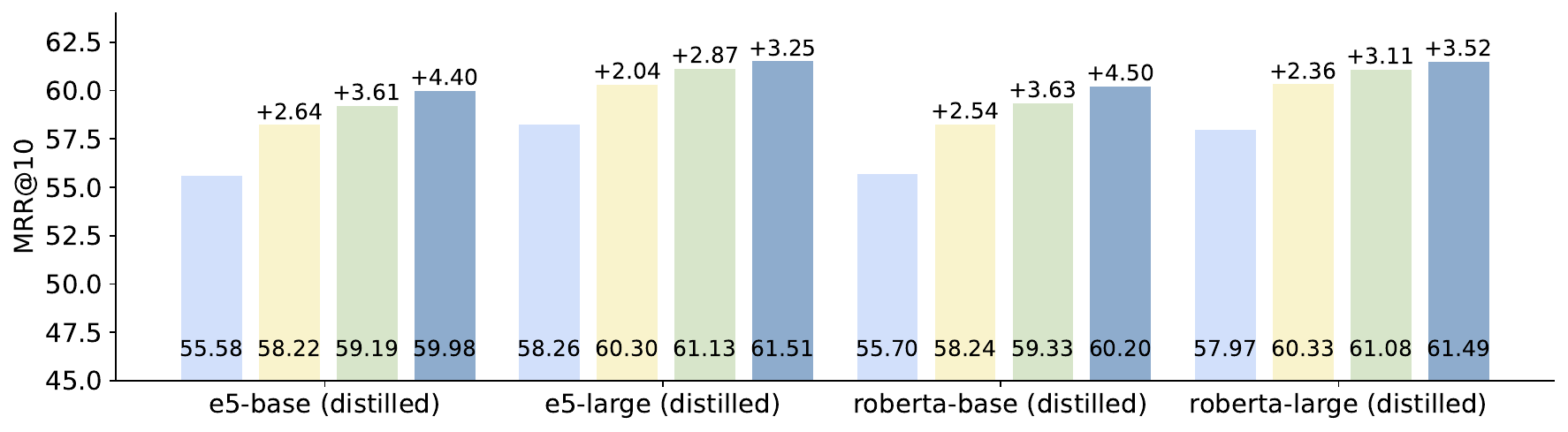}
\caption{MRR@10 values obtained by the models trained in our study. We present the results of the original multilingual E5 models, as well as models distilled from FlagEmbeddings based on E5 and Polish RoBERTa. For each model, we show its performance before fine-tuning, after fine-tuning on the Polish MS MARCO dataset, and the result of the sparse-dense hybrid combining the given model with BM25 and SPLADE methods.}
\label{fig:mrr}
\end{figure*}

\begin{figure*}[htp]
\includegraphics[scale=0.53]{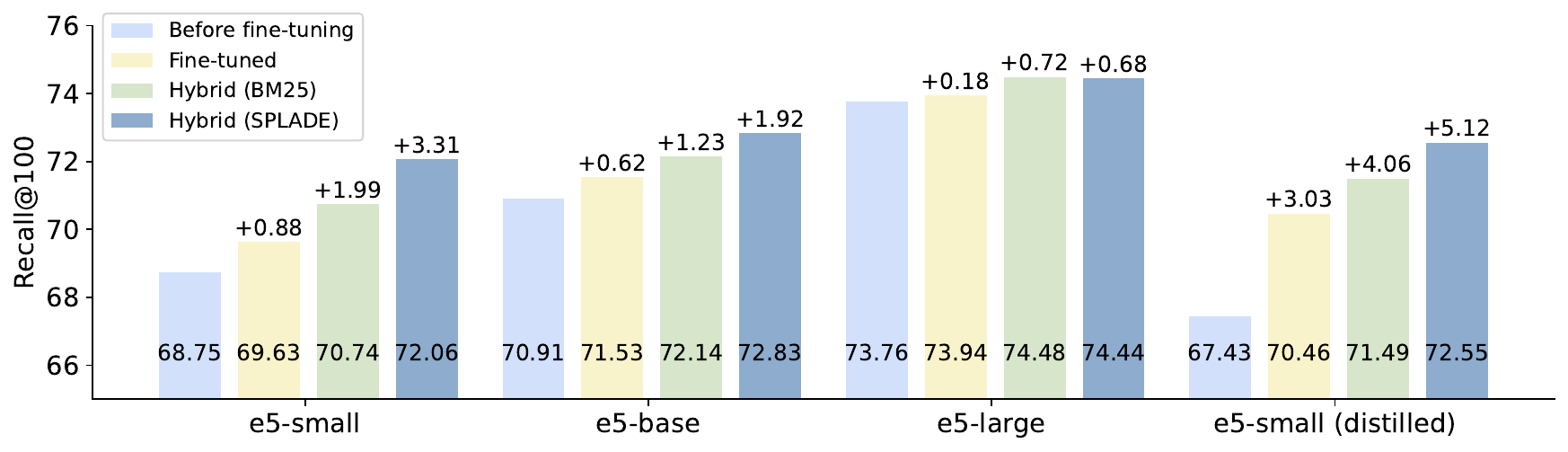}\\
\includegraphics[scale=0.53]{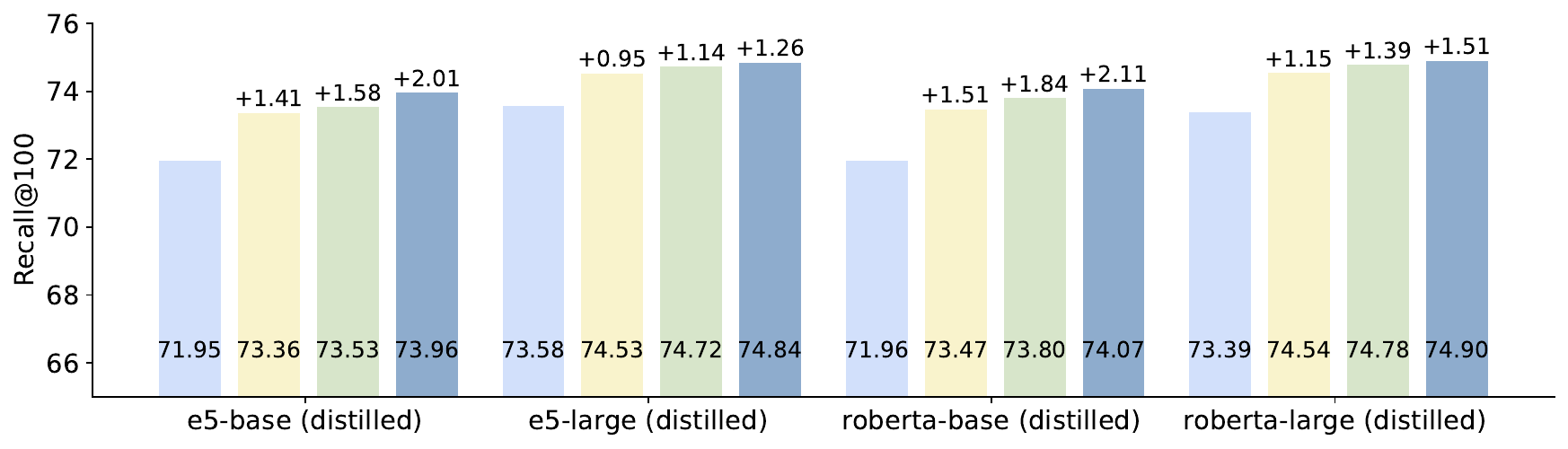}
\caption{Recall@100 values obtained by the models trained in our study. We present the results of the original multilingual E5 models, as well as models distilled from FlagEmbeddings based on E5 and Polish RoBERTa. For each model, we show its performance before fine-tuning, after fine-tuning on the Polish MS MARCO dataset, and the result of the sparse-dense hybrid combining the given model with BM25 and SPLADE methods.}
\label{fig:recall}
\end{figure*}

\begin{figure*}[htp]
\includegraphics[scale=0.53]{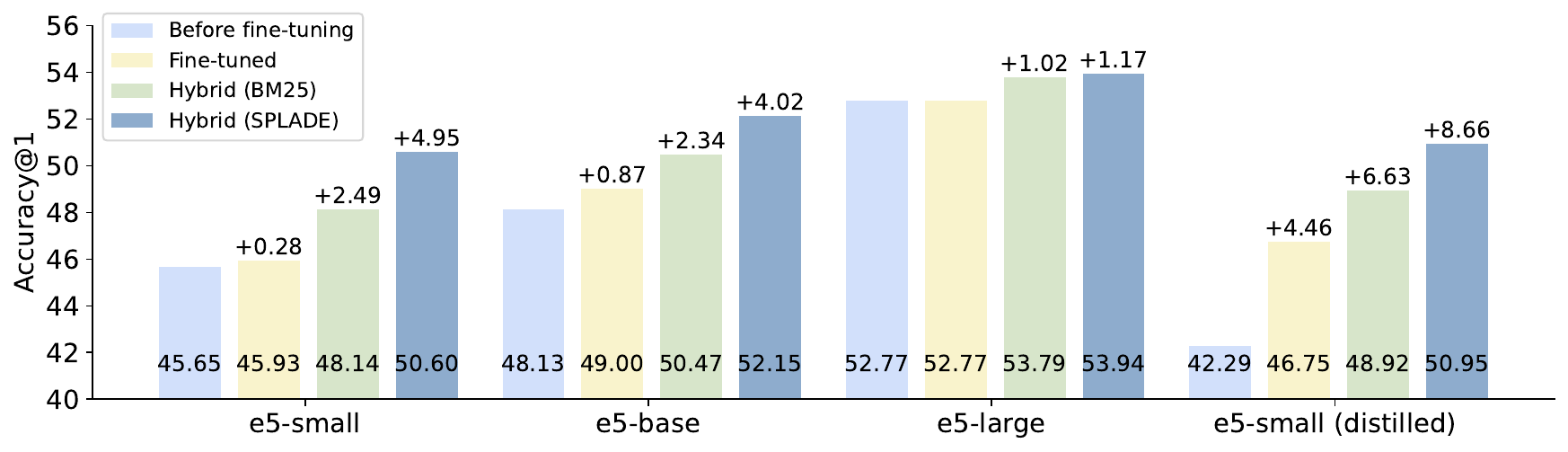}\\
\includegraphics[scale=0.53]{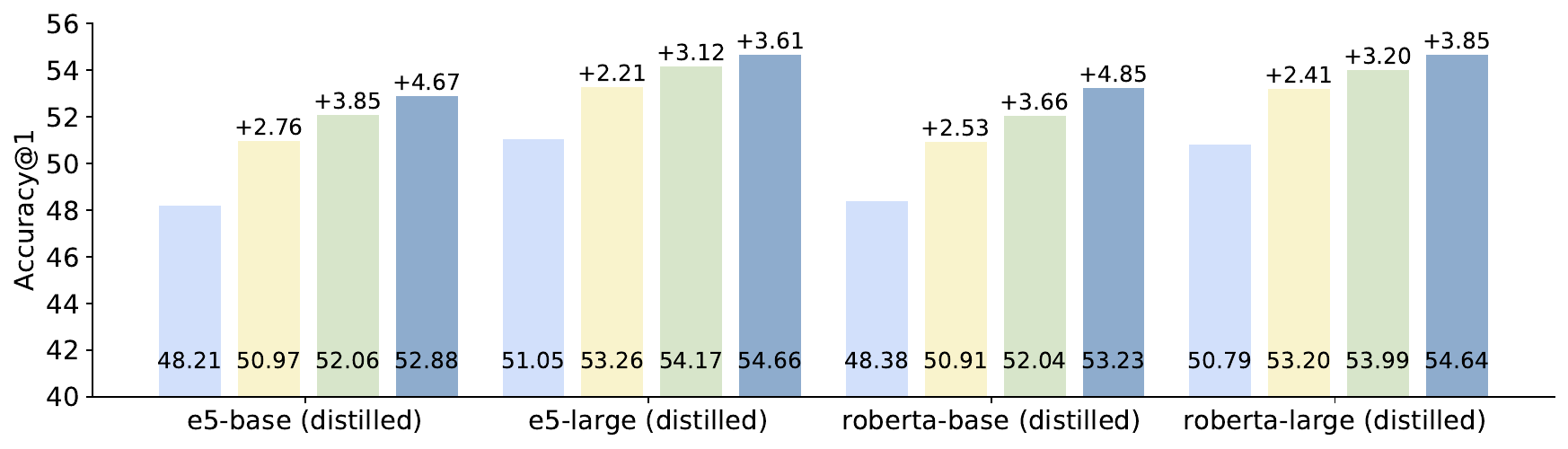}
\caption{Accuracy@1 values obtained by the models trained in our study. We present the results of the original multilingual E5 models, as well as models distilled from FlagEmbeddings based on E5 and Polish RoBERTa. For each model, we show its performance before fine-tuning, after fine-tuning on the Polish MS MARCO dataset, and the result of the sparse-dense hybrid combining the given model with BM25 and SPLADE methods.}
\label{fig:acc}
\end{figure*}

\end{document}